\documentclass{article}

\usepackage[english]{babel}

\usepackage[letterpaper,top=2cm,bottom=2cm,left=3cm,right=3cm,marginparwidth=1.75cm]{geometry}

\usepackage{amsmath}
\usepackage{amsfonts}
\usepackage{amssymb}
\usepackage{graphicx}
\usepackage{subfigure}
\usepackage[colorlinks=true, allcolors=blue]{hyperref}
\usepackage{comment}
\usepackage{float}
\usepackage{booktabs} 
\usepackage{enumerate}
\usepackage{apacite}
\usepackage{color}
\usepackage {appendix}

\title{Double Deep Q-Learning in Opponent Modeling}
\author{Yangtianze Tao \thanks{Department of Mathematical Sciences, Tsinghua University, Beijing 100084, China 
{(E-mail: tytz19@mails.tsinghua.edu.cn).}}
\ and John Doe
}

\begin{document}
\maketitle

\begin{abstract}
Multi-agent systems in which secondary agents with conflicting agendas also alter their methods need opponent modeling. In this study, we simulate the main agent's and secondary agents' tactics using Double Deep Q-Networks (DDQN) with a prioritized experience replay mechanism. Then, under the opponent modeling setup, a Mixture-of-Experts architecture is used to identify various opponent strategy patterns. Finally, we analyze our models in two environments with several agents. The findings indicate that the Mixture-of-Experts model, which is based on opponent modeling, performs better than DDQN.

\end{abstract}

\section{Introduction}

Recently, multi-agent systems have become more important in the fields of computer science and civil engineering.  \cite{dorri2018multi}. Recent studies have concentrated their attention on reinforcement learning as an efficient control approach for robot systems, autonomous driving, resource utilization, and other related topics. \cite{du2021survey}. When there are multiple agents present in a place, the environment is affected by the sum of the activities of all of the actors. Because of this, for agents to engage in single-agent reinforcement learning, they are required to interact with other agents, and their returns are contingent on the actions of other agents. One may argue that the opponents are just an aspect of the total environment. However, seeing opponents in this way might impede reactions to adaptive opponents  \cite{Pinto2017RobustAR}, since the change in behavior is obscured by the world's dynamics. The opponent modeling for reinforcement learning \cite{he2016opponent} is a helpful framework for a single agent learning in a multi-agent environment, allowing the agent to exploit the policies and other important information of different opponents. 

In this study, we concentrate on the Double Deep Q-learning architecture and examine the effects of the various opponent policies on the learning of the main agent inside the opponent modeling framework. First, we establish a robust secondary agent policy by competing against one another. Then, as the secondary agent, we designate this technique as our adversary. Then, we retrain a main agent utilizing the opponent characteristics retrieved by the Mixture-of-Experts architecture against this opponent.

The structure of this paper is as follows: In Section 2, the reinforcement learning (RL) context is reviewed. Then, the traditional and deep reinforcement learning techniques are briefly presented. Section 3 contains the structure of the opponent modeling for deep Q-learning as well as the Mixture-of-Experts architecture for our models. Afterwards, Section 4 contains details on the experimental design and outcomes of this study, as well as a discussion of the results. Section 5 presents the findings.

\section{Reinforcement Learning Prelimiaries}
{\color{black}
The mathematical basis of reinforcement learning is Markov Decision Processes (MDPs) \cite{sutton2018reinforcement}. An MDP usually consists of a state space, an action space, a state transition matrix, a reward function, and a discount factor. Simply put, reinforcement learning is a sequential decision-making process that tries to find a decision rule that makes the system get the maximum cumulative reward value, i.e. get the maximum value.

The states and actions of the agent are defined by the state space $\mathcal{S}$ and the action space $\mathcal{A}$. The state transition function defined by
\begin{equation}
\mathcal{P}_{ss^{\prime}}^{a}=\mathbb{P}\left(S^{\prime}=s^{\prime} \mid S=s, A=a\right),
\end{equation}
which is the transition probability of the agent takes an action $a$ in state $s$. The agent's behavior is defined by a policy $\pi$ such that $\pi(a \mid s)=\mathbb{P}(A=a\mid S=s)$ is the probability of taking action $a$ in state $s$. After the agent takes a action, it will receive a rewards $r$. Rewards $r$ are assigned by a real-valued reward function $\mathcal{R}\left(s, a, s^{\prime}\right)$, which is a value returned by the environment to the agent after the agent performs an action. The reward is often defined by ourselves. It is necessary to note that how well the reward is defined greatly affects the outcome of reinforcement learning.

We define the discounted cumulative return at time step $t$, $U_{t}=R_{t}+\gamma R_{t+1} + \gamma^{2}R_{t+2}+\dots $, where $\gamma \in[0,1]$ is the discount factor. The purpose of reinforcement learning is to maximize the the expected discounted cumulative return $R=\mathbb{E}\left[U_{t}\right]$. 

\subsection{Two RL pattern}

Recall the goal of the agent is to maximize the discounted return, that is, maximize the weighted sum of the rewards. Once the goal is defined, reinforcement learning methods can be designed to achieve the goal. Reinforcement learning methods are generally divided into two categories: model-based methods and model-free methods, and we focus on the latter. Model-free methods can be further divided into value learning and policy learning.

Here, we shall first introduce the action-value function and the state-value function based on return $U_{t}$. The action-value function for policy $\pi$ and optimal action-value function are defined as follow:
\begin{equation}\label{eqn:action-value}
\begin{aligned}
    Q_{\pi}(s_{t}, a_{t}) & = \mathbb{E}\left[U_{t}\mid S_{t}=s_{t}, A_{t}=a_{t}\right] \\
    Q^{\star}(s_{t}, a_{t}) & = \max_{\pi} Q_{\pi}(s_{t}, a_{t}).
\end{aligned}
\end{equation}
Notice that  $U_{t}=\sum_{k=t}^{n}\gamma^{k-t}R_{k}$ and $U_{t+1}=\sum_{k=t+1}^{n}\gamma^{k-t-1}R_{k}$, which implies
\begin{equation}
    U_{t} = R_{t} + \gamma\sum_{k=t+1}^{n}\gamma^{k-t-1}R_{k}.
\end{equation}
Then combine \eqref{eqn:action-value}, we can conclude that the well-known optimal bellman equations:
\begin{equation}\label{eqn:bellman-eq}
        Q^{\star}(s_{t}, a_{t}) = \mathbb{E}_{S_{t+1}\sim p(\cdot\mid s_{t}, a_{t})}\left[R_{t}+\gamma\max_{A\in\mathcal{A}}Q^{\star}(S_{t+1}, A)\mid S_{t}=s_{t}, A_{t}=a_{t}\right].
\end{equation}

The state-value function is defined as follows:
\begin{equation}\label{eqn:state-value}
    V_{\pi}(s_{t})=\mathbb{E}_{A}\left[Q_{\pi}(s, A)\right]
\end{equation}
The agent can be controlled by either $\pi(a\mid s)$ or $Q^{\star}(s, a)$, which indicates the Value-Based learning and Policy-Based learning.

Value-Based learning usually refers to learning the optimal value function $Q^{\star}(s, a)$. If we have $Q^{\star}$, the agent can make decisions based on $Q^{\star}$ and choose the best action. Each time a state $s_{t}$ is observed, feed it into the $Q^{\star}$ and let $Q^{\star}$ evaluate all actions. Then the agent's decision can be represented by this formula:
\begin{equation}
    a_{t} = \arg\max_{a\in\mathcal{A}}Q^{\star}(s_{t}, a)
\end{equation}

Policy-Based learning refers to learning the policy function $\pi(a\mid s)$. If we have a policy function, we can directly use it to calculate the probability values of all actions, then randomly sample an action and execute it. Each time a state $s_{t}$ is observed, feed it into the $\pi$, let $\pi$ evaluate all actions, and get the probability value.

 Mathematically, we must dynamically solve the optimal bellman equations \eqref{eqn:bellman-eq}. However, there is no closed form solution. Consequently, several iterative solutions are offered. And in this paper, we are focus on the Value-Based learning method.

}
\subsection{Classic Reinforcement Learning Methods}

\subsubsection{Q-learning}

Here we introduce the well-known Q-learning \cite{watkins1992q}. The optimal policy for a given state always selects the action with the greatest Q-value. Q-learning use Q-Table to store the Q value associated with each state-action combination.

Assume that both the state space $\mathcal{S}$ and the action space $\mathcal{A}$ are finite sets, i.e., they have a limited number of elements. For instance, S has three states and $\mathcal{A}$ contains $4$ actions. Then the optimal action-value function $Q^{\star}(s, a)$ can be represented as a $3\times 4$ table. The formula used when making decisions based on the current state $s_{t}$:
\begin{equation}
    a_{t} = \arg\max_{a\in\mathcal{A}}Q^{\star}(s_{t}, a).
\end{equation}

The main idea of the Q-learning is to use a table $\Tilde{Q}$ to approximate $Q^{\star}$. It can summarised as follows:

\begin{enumerate}[i)]
    \item First initialize $\Tilde{Q}$. It can be a table of all zeros.
    \item $\Tilde{Q}$ is then updated using a tabular Q-learning algorithm, one element of the table at a time. 
    \item In the end, $\Tilde{Q}$ will converge to $Q^{\star}$.
\end{enumerate}
Now recall the optimal Bellman equation \eqref{eqn:bellman-eq}, we approximate the left and right sides of the Eq \eqref{eqn:bellman-eq}:
\begin{itemize}
    \item $Q^{\star}(s_{t}, a_{t})$ on the left-hand side of the equation can be approximated as $\Tilde{Q}(s_{t}, a_{t})$, where $\Tilde{Q}(s_{t}, a_{t})$ is the table's estimate of $Q^{\star}(s_{t}, a_{t})$ at time t.
    \item The expectation on the right side of the equation is found with respect to the next moment state $S_{t+1}$. Given the current state $s_{t}$, the agent performs the action $a_{t}$, and the environment gives a reward $r_{t}$ and a new state $s_{t+1}$. A Monte Carlo approximation to the expectation using the observed $r_{t}$ and $s_{t+1}$ yields
    \begin{equation}\label{eqn:q-learning-1}
        r_{t} + \gamma\max_{a\in\mathcal{A}}Q^{\star}(s_{t+1}, a).
    \end{equation}
\end{itemize}

We further approximate $Q^{\star}$ by $\Tilde{Q}$ in \eqref{eqn:q-learning-1}. It implies
\begin{equation}
    \hat{y}_{t}=r_{t} + \gamma\max_{a\in\mathcal{A}}\Tilde{Q}(s_{t+1}, a),
\end{equation}
which is called the TD target. It is the table's estimate of $Q^{\star}(s_{t}, a_{t})$ at time $t + 1$.

Both $\Tilde{Q}(s_{t}, a_{t})$ and $\hat{y}_{t}$ are estimates of the optimal action value $Q^{\star}(s_{t}, a_{t})$. Since $\hat{y}_{t}$ is partly based on real observed awards $r_{t}$. We think that $\hat{y}_{t}$ can be served as the correction for the $\Tilde{Q}(s_{t}, a_{t})$. So, we update $\Tilde{Q}(s_{t}, a_{t})$ as follows:
\begin{equation}\label{eqn:q-learning-update}
    \Tilde{Q}(s_{t}, a_{t}) \longleftarrow (1-\alpha)\Tilde{Q}(s_{t}, a_{t}) + \alpha\hat{y}_{t},
\end{equation}
{\color{black}where $\alpha$ is called learning rate or step size, which is usually a small integer that determines the contribution of each correction based on $\hat{y}_{t}$. \eqref{eqn:q-learning-update} makes $\Tilde{Q}(s_{t}, a_{t})$ closer to $\hat{y}_{t}$. The goal of Q-learning is to make $\Tilde{Q}$ gradually approach $Q^{\star}$.

Now, we have introduced the general framework of Q-learning. Next, we cover some of the implementation details, namely how to collect training data. We hope that the agent's behavior strategy needs to be accompanied by some randomness, that is, each time there is a small probability to explore the behavior that is not the most strategic at present. This strategy for controlling the agent is called behavior policy, and the more commonly used behavior policy is $\epsilon-$greedy:

\begin{equation}
    a_{t} = \begin{cases}
 \arg\max_{a}\Tilde{Q}(s_{t}, a) & \text{with probability } 1 - \epsilon\\
\text{Randomly draw an action in } \mathcal{A} & \text{with probability} \epsilon.
\end{cases}
\end{equation}
}

\subsection{Deep Reinforcement Learning Methods}
{\color{black}
In this previous section, we have introduced the formulation of Q-learning. When the state and action space are discrete and the dimension is low, the aforementioned Q-learning technique is successful.  However, if the state and action space is high-dimensional and continuous, it will suffer from  "curse of dimensionality", in which the amount of processing rises exponentially as the number of dimensions increases. Incorporating deep neural networks into the RL begin with \cite{mnih2013playing}. The idea is to use a deep neural network to approximate optimal value function. This network is called deep Q-network (DQN). Here we shall introduce the DQN algorithm and its modification called Double deep Q-network (DDQN) \cite{van2016deep}.
\subsubsection{Deep Q-Learning}
DQN use a deep neural network $Q(s, a;\theta)$ to approximate optimal value function $Q^{\star}(s, a)$ using an experience replay mechanism. We shall explain what is experience replay mechanism and why it is necessary later. Here $Q(s, a;\theta)$ usually refer to the target network and $\theta$ is the network's trainable parameters for target network. And we also call the network to approximate the value function as the Q-network.

The motivation why using deep neural networks is twofold. It is well-known that the feed-forward neural networks are the universal approximator for the measurable function \cite{HORNIK1990551}. The second one is that why we need the deep networks and the Universal Approximation Theorem already tells me that a two-layer neural network can approximate any measurable function. This motivation can be found in in \cite{Liang2017WhyDN}, which theoretically proves that in order to achieve the same fitting error, the number of neurons required by a deeper neural network is much smaller than that of a neural network with fewer layers. Back to DQN, after action $a_{t}$ is taken in state $s_{t}$ and observing the immediate reward $r_{t+1}$ and resulting state $s_{t+1}$ is then
 \begin{equation}
     \theta_{t+1} = \theta_{t} + \alpha\left(Y_{t}^{\text{Q}} - Q(s_{t}, a_{t};\theta_{t})\right)\nabla_{\theta_{t}}Q(s_{t}, a_{t};\theta_{t}),
 \end{equation}
 where $\nabla_{\theta_{t}}$ denotes partial derivative operator with respect to $\theta_{t}$ and $\alpha$ is a scalar step size and the target $Y^{\text{Q}}_{t}$ is defined as follows:
 \begin{equation}
     Y^{\text{Q}}_{t} := r_{t+1} + \gamma\max_{a}Q(s_{t+1}, a;\theta_{t}).
 \end{equation}
 }

\subsubsection{Double Deep Q-Learning}

Since both Q-learning and DQN algorithms suffer from overestimation problem, i.e., they overestimate the Q value, resulting in overoptimistic estimates values. Double Q-learning \cite{hasselt2010double} has been proposed as a solution to this issue, and has been expanded to a new reinforcement learning technique known as deep double Q Network (DDQN) \cite{van2016deep}.

{\color{black}The main idea of DDQN is to introduce two Q-network. One is called target network, which corresponds to the target network in DQN. The other one is called evaluation network. The motivation is based on the double Q-learning. When given the current state, we choose an action based on the value of the target network. Since we have mentioned above, this step is the original DQN suffering from the overestimation problem. In the Double Q-Learning algorithm, the extra step is that the evaluation network needs to evaluate this action. Because the parameters of the two Q-networks are different, the evaluation of the same action will also be slightly different. We choose the smaller value evaluated to choose the action. This avoids overestimate problem.

We denote $\theta$ and $\theta^{\prime}$ as the parameters for the target network and evaluation network, respectively.} Then the Double Q-learning error can then be expressed as follows:
\begin{equation}
    Y_{t}^{\text{DoubleQ}}:= r_{t+1} + \gamma Q\left(s_{t+1}, \arg\max_{a}Q(s_{t+1}, a;\theta_{t});\theta_{t}^{\prime} \right).
\end{equation}
This indicates that we continue to make use of the greedy policy in order to learn how to estimate the value of Q, and that we make use of the second weight parameters $\theta^{\prime}$ in order to assess the policy's effectiveness.
 
 \subsubsection{Experience Replay and Prioritized Experience Replay}
 {\color{black}
 At the end of this part, we shall introduce the experience replay mechanism \cite{lin1992self}, which is an important technique in RL. It can greatly improve the performance of RL algorithm. Then we introduce the Prioritized Experience Replay (PER) \cite{schaul2015prioritized}, which is modification the original experience replay. And this technique will be used in our implementation.

Experience replay means storing the records of the agent's interaction with the environment into an array, and using these experiences repeatedly to train the agent afterwards. This array is called Replay Buffer. More specifically, We can control the agent with any strategy, interact with the environment, and divide the obtained trajectory into quadruplet such as $e_{t} = (s_{t}, a_{t}, r_{t}, s_{t+1})$, also known as experience transition. And store them in the Replay Buffer denoted by $E$. After every action that is taken in the environment, add the new experience transition $e$ into the set $E$. $E$ has a certain predefined capacity, where upon reaching its maximal size, the oldest experience transitions are continually replaced with newer ones. In this way, the experience replay is similar to a queue, which is a commonly used data structure in algorithm implementation.

A natural question is why we need experience replay mechanism in the agent learning process. Generally speaking, the agent will forget about previously-occurring events. With the inclusion of experience replay, the agent is able to learn from events of the past information. Specifically, there are the following two benefits:

\begin{enumerate}[(i)]
    \item One benefit of experience replay is to break down sequence dependencies. When training the DQN, each time we update the parameters of the DQN with a quadruplet. We want the quadruplets used twice adjacent to be independent. However, when the agent collects experience, the two adjacent quadruplets $(s_{t}, a_{t}, r_{t}, s_{t+1})$ and $(s_{t+1}, a_{t+1}, r_{t+1}, s_{t+2})$ are strongly correlated. Training a DQN using these strongly correlated quadruplets in sequence tends to perform poorly. The experience playback randomly selects a quadruplet from the array each time, which is used to update the DQN parameters once. In this way, the randomly drawn the quadruplets are all independent, eliminating the correlation.
    \item Another benefit of experience replay is that the collected experience is reused rather than discarded once, so that the same performance can be achieved with a smaller number of samples.
\end{enumerate}
Hence, experience replay mechanism allows the agent to both remember positive actions made in specific contexts and to learn from bad action choice.

After introducing the experience replay, we may notice that the experience replay mechanism was equally sampled from a Replay Buffer. In other word, only repeats transitions at the same frequency with which they were first seen, regardless of their significance. Based on this intuition, prioritized experience replay (PER) is proposed, which is a special experience replay method that works better than original experience replay with both faster convergence and higher average returns upon convergence.

The main idea of the PER is to give each quadruplet a weight, and then do non-uniform random sampling according to the weight. If the value judgment of $(s_{t}, a_{t})$ by DQN is inaccurate, that is, $Q(s_{t}, a_{t};\theta)$ is far from $Q^{\star}(s_{t}, a_{t})$, then the quadruplet $(a_{t}, s_{t}, r_{t}, s_{t+1})$ should be have higher weight.

Since $|Q(s_{t}, a_{t};\theta) -Q^{\star}(s_{t}, a_{t})|$ can not be obtained, we consider the temporal difference (TD) error:
\begin{equation}\label{eqn:td}
    \delta_{t} = Q(s_{t}, a_{t};\theta) - \left[r_{t} + \gamma \max_{a\in\mathcal{A}}Q(s_{t+1}, a;\theta)\right].
\end{equation}
If the absolute value of the TD error $|\delta_{t}|$ is large, it means that the current DQN is inaccurate in evaluating the true value of $(a_{t}, s_{t})$, then it should give the high weight for $(a_{t}, s_{t}, r_{t}, s_{t+1})$.

PER will do the non-uniformly sampling in the array. The weight of the quadruple $(a_{t}, s_{t}, r_{t}, s_{t+1})$ is the absolute value of the TD error, and its sampling probability depends on the TD error. The sampling probability $\mathbb{P}_{t}$ is usually set as follows:
\begin{equation}
    \mathbb{P}_{t}\propto |\delta_{t}| + \epsilon,
\end{equation}
where $\epsilon$ is a small number that prevents the sampling probability from approaching zero, which is used to ensure that all samples are drawn with a non-zero probability.

}

\section{Opponent Modeling}

When there are multiple agents present in the same environment, the environment will be affected by
the joint action of all agents. And the consequence of one action based on a specific state is no longer stable from the perspective of a single agent, but it is influenced by the actions of other agents.

In MDP terms, the joint action space is defined by $A_{M}=A_{1}\times A_{2}\times\dots\times A_{n}$, where $n$ is the total number of agents. We use $A$ to denote the action of the agent we
control (the primary agent) and $O$ to indicate the joint action of all other agents (secondary agents), such that $(A, O)\in A_{M}$. Let $\mathcal{T}^{M}(s, a, o, s^{\prime})=\mathbb{P}(S^{\prime}=s^{\prime}\mid S=s, A=a, O=o)$ to denote the state transition function, and {\color{black}= $\mathcal{R}^{M}(s, a, o, s^{\prime})$ to denote the new reward function. The goal of introducing the new reward function is that the primary agent and secondary agent interact with the environment together under the multi-agent environment}. The purpose of reinforcement learning in the multi-agent setting is to determine, given the interactions of the main agent with the combined policies of the secondary agents, which policy will provide the greatest results for the main agent.

\subsection{Opponent Modeling for Q-Learning}

The optimal Q-function relative to the joint policy of opponents: $Q^{\star\mid\pi^{o}}=\max_{\pi}Q^{\pi\mid\pi^{o}}(s, a)$ for all $s\in\mathcal{S}$ and all $a\in\mathcal{A}$. And the recurrent relation between Q-values holds:
\begin{equation}\label{eqn:opponent-model}
\begin{aligned}
    Q^{\pi\mid\pi^{o}}(s_{t}, a_{t}) & = \sum_{o_{t}}\pi_{t}^{o}\sum_{s_{t+1}}\mathcal{T}(s_{t}, a_{t}, o_{t}, s_{t+1})\left[\mathcal{R}(s_{t}, a_{t}, o_{t}, s_{t+1}) + \gamma\mathbb{E}_{a_{t+1}}\left[Q^{\pi\mid \pi^{o}}(s_{t+1}, a_{t+1})\right]\right] \\
    & = \sum_{o_{t}}\pi_{t}^{o}(o_{t}\mid s_{t})Q^{\pi}(s_{t}, a_{t}, o_{t}).
\end{aligned}
\end{equation}

The deep Reinforcement Opponent Network models $Q^{\cdot\mid\pi^{o}}$ and $\pi^{o}$ jointly, which consists of a Q-Network $N_{Q}$ that
evaluates actions for a state and an opponent network $N_{o}$
that learns representation of $\pi^{o}$.

\subsection{Mixture-of-Experts for Deep Q-Learing}
{\color{black}
We shall review some preliminaries of the neural networks. The neural networks model we used in the paper are feed-forward neural networks with two hidden layer. And the activation function is Rectified Linear Units (ReLU) \cite{Agarap2018DeepLU}. For the input $x\in\mathbb{R}^{d}$ and output $y\in\mathbb{R}^{r}$, this neural networks can be formulated as the follows:
\begin{equation}\label{eqn:relu-network}
    y = W_{2}\cdot\text{ReLU}(W_{1}x+b_{1}) + b_{2},
\end{equation}
where $W_{2}\in\mathbb{R}^{r\times h}, W_{1}\in\mathbb{R}^{h\times d}, b_{1}\in\mathbb{R}^{h}, b_{2}\in\mathbb{R}^{r}$ are the trainable parameters. And the ReLU function is defined by
\begin{equation}\label{eqn:relu}
    \text{ReLU}(x) = \max(0, x), \quad, x\in\mathbb{R}.
\end{equation}

In theory, if we wish to model all strategies of the opponents, we need to explicitly compute \eqref{eqn:opponent-model}. However, this is difficult to do so in most cases. We propose an alternative approach called double deep Q-network with Mixture-of-Experts (DDQN-MOE) to model the opponent's strategies adaptively. This model consists of three neural networks. Let $\phi^{s}$ and $\phi^{o}$ represent the agent's and opponent's respective states. The first neural network model embed $\phi^{s}$ and $\phi^{o}$ to $h^{s}$ and $h^{o}$ as hidden states. This network can be served as the  features extractor. The remaining two neural network models are $K$ parallel DDQNs $Q_{i}$ and Mixture-of-Experts (MOE) \cite{6797059} network. For $i=1,2, \dots, K$, $Q_{i}$ is defined by
\begin{equation}\label{eqn:ddqn-seq}
    Q_{i}(h^{s}, \cdot) = f(W_{i}^{s}h^{s} + b_{i}^{s}).
\end{equation}
These $Q_{i}$ encode the information of $h^{s}$ into the DDQN. Then MOE network needs to explicitly represent the opponent action as a hidden state and marginalize it.  The expected Q-value is obtained by combining predictions from multiple expert networks:
\begin{equation}\label{eqn:moe}
\begin{aligned}
    Q(s_{t}, a_{t};\theta)  = \sum_{i=1}^{K}\omega_{i}Q_{i}(h^{s}, a_{t}). 
\end{aligned}
\end{equation}
Here the opponent representation combination weights $\omega$ is computed by
\begin{equation}\label{eqn:opponent-weight}
    \omega = \text{softmax}(f(W^{o}h^{o}+b^{o})),
\end{equation}
where the value of the softmax function for $x=(x_{1}, x_{2}, \dots, x_{n})\in\mathbb{R}^{n}$ defined as follows:
\begin{equation}
    \text{softmax}(x) = (\frac{\exp(x_{1})}{\sum_{i=1}^{n}\exp(x_{i})}, \frac{\exp(x_{2})}{\sum_{i=1}^{n}\exp(x_{i})}, \dots, \frac{\exp(x_{n})}{\sum_{i=1}^{n}\exp(x_{i})}).
\end{equation}

The schematic structure of DDQN-MOE is shown in Fig \ref{fig:MOE}.

\begin{figure}[H]
    \centering
    \includegraphics[scale=0.5]{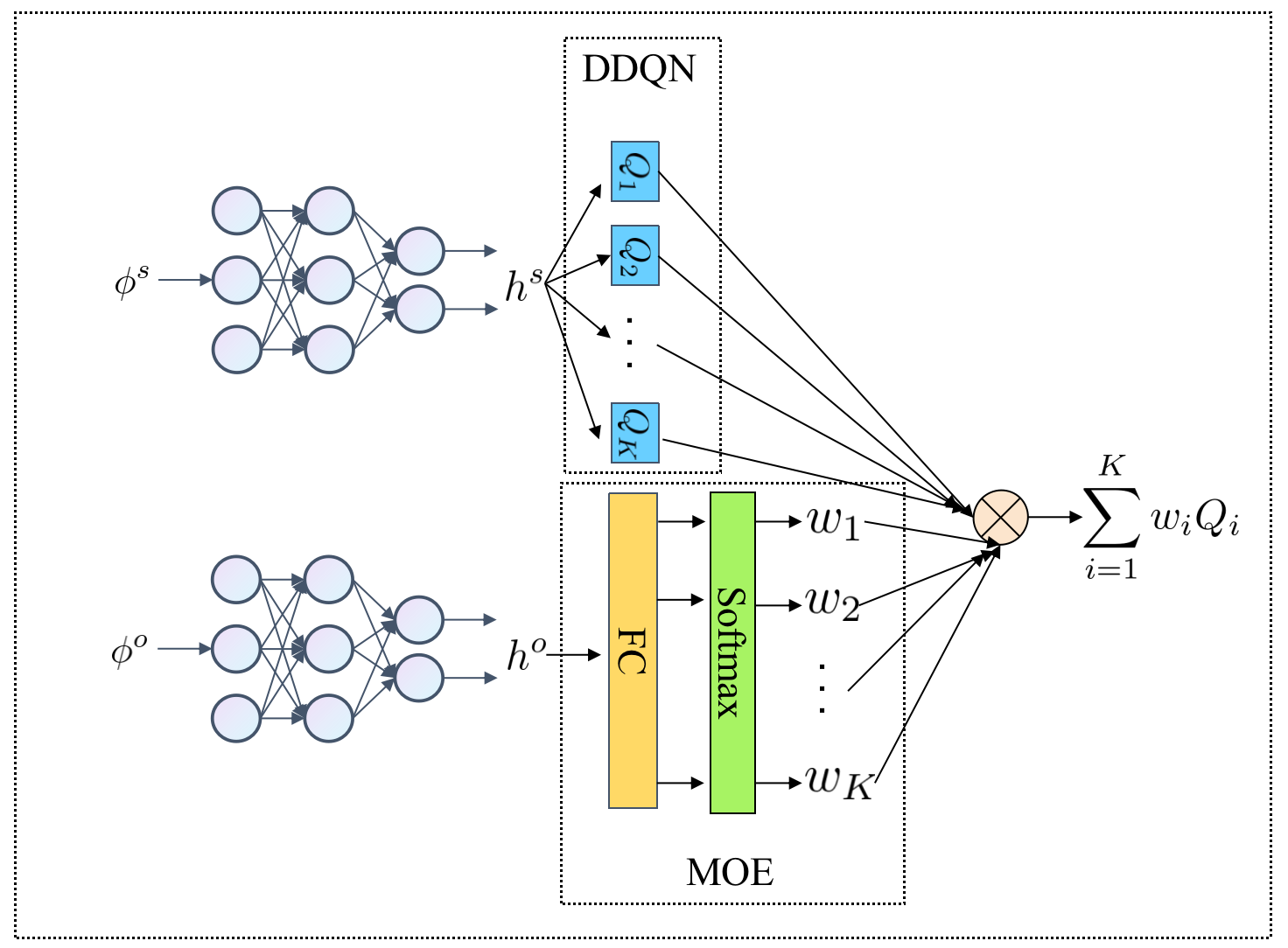}
    \caption{The schematic structure of DDQN-MOE. Here the FC represents the fully connected layer in the neural network, which follows \eqref{eqn:opponent-weight}.}
    \label{fig:MOE}
\end{figure}

It needs to note that the MOE network disregards interaction between the environment and opponent behavior, i.e., $w_{i}$ depends only on $h^{o}$, not on $h^{s}$. The motivation is noticing that Q-values have various distributions based on $\phi^{o}$ and that each expert network record a single type of opponent strategy. 

}

\section{Experiments}

\subsection{Experimental Setup}

In this section, we shall evaluate our models on two tasks from the Multi Particle Environments (MPE) \cite{lowe2017multi}, which are a set of communication oriented environment where particle agents can (sometimes) move, communicate, see each other, push each other around, and interact with fixed landmarks. We use the implementation in PettingZoo \cite{terry2021pettingzoo}.

{\color{black}
\subsubsection{The opponent strategy}
\label{sec:opp-str}
To investigate opponent modeling for these two tasks, we first need to get an opponent with a certain effective strategy. {\color{black}The usual approach is to artificially devise some rule-based tricks, like the football game in the \cite{he2016opponent}. They design two-mode rule-based agent as the opponent. But for the two games we've chosen, developing effective rules-based tricks is difficult. The reason is that the football game has relatively clear rules of victory and defeat, and its state and action space are discrete. But the game we chose was much more complex, with a continuous state space and no very clear rules for winning or losing.
 Therefore, we use another method to get a powerful opponent agent. We aim to obtain a powerful opponent agent through randomly initializing two agents one as primary agent and the other be secondary agent, and let them learn against each other. They will learn from each other in enough episodes and explore each other's strategies. We call this approach adversarial learning initialization (ALI). This idea is like to the Adversarial Generative Networks (GAN) \cite{goodfellow2014generative}, eventually with powerful agents and powerful adversary agent. Then we can continue to study our proposed model based on them. The result of ALI is that each agent trained via ALI is the powerful opponent against the other one agent with random initialized strategy. 
 
 \subsubsection{The opponent features}
 \label{sec:opp-features}
 After introducing how to create the opponent strategy, we proceeding with how to design the opponent features. In these two task, they share the similar agents setting, i.e., there are two agents, one is called good agent and the other one is called adversary. And their action space are all defined by five cases: no action, move left, move right, move down and move up. So their action is a $1\times 5$ vector. Motivated by the well-known Alpha-Go \cite{Silver2017MasteringTG}, which beats the world champion at the task of Go. In their algorithm design, opponent modeling is also considered, which specifically takes the opponent's past 7 moves as features for opponent modeling. We introduce opponent features as the normalized frequencies of the observed opponent actions and its last action. Adding the last action into the opponent features has been verified in \cite{he2016opponent}. Therefore, the opponent feature is a $1\times 6$ vector. And the normalized frequencies is computed as follows:
\begin{equation}\label{eqn:opponent-features}
    p_{\text{norm}} = [\frac{N_{1}}{N}, \frac{N_{2}}{N}, \frac{N_{3}}{N}, \frac{N_{4}}{N}, \frac{N_{5}}{N}],
\end{equation}
where $N_{1}, \dots, N_{5}$ and $N$ are the number of the every observed opponent actions and total number of the all observed actions. 
}

\subsubsection{Experimental Design}
Here we shall introduce the specific experimental process. All the methods used in this paper are explained in detail within previous section. And all experimental code implementations in this paper are based on Pytorch \cite{Paszke2019PyTorchAI} and Numpy \cite{Harris2020ArrayPW}. The parameter configurations used for each of the methods will be listed in Appendix \ref{appendix-params}.

 For the above two tasks, we used the same training and testing patterns. More specifically, it is divided into the following two steps:

\begin{itemize}
    \item \textbf{Obtaining powerful opponent strategies}: This step has already been mentioned above. We randomly initialize the agent and adversary, respectively, and then let them play until they reach a high level of equilibrium that restricts each other. In this way, one of the trained agents is a powerful opponent relative to another randomly initialized model.
    \item \textbf{learning against the powerful opponent model}: In the previous step, after we selected which agent to use as the opponent model, we kept its trained parameters fixed and let it learn by playing against another re-randomly initialized agent using opponent features and MOE model.
\end{itemize}

 The purpose of introducing adversary modeling is to explore whether the information of the adversary can break the equilibrium we established through ALI, thus serving as an effective learning paradigm in multi-agent systems.
}

\subsection{Simple Push}

\begin{figure}[H]
    \centering
    \includegraphics[scale=0.5]{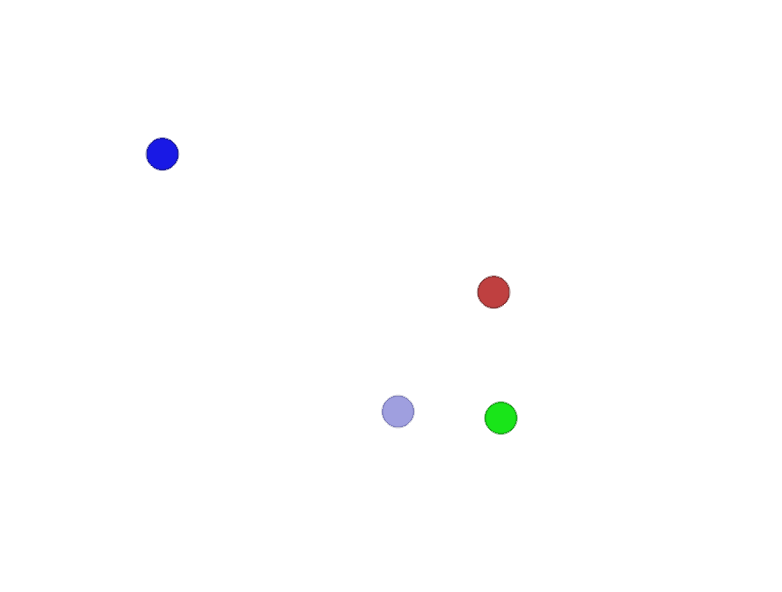}
    \caption{Random Policy on Simple Push}
    \label{fig:random-simple-push}
\end{figure}

This environment has $1$ good agent (green), $1$ adversary (red), $1$ goal landmark and $1$ additional landmark. An example of a random policy is illustrated as Fig \ref{fig:random-simple-push}.

The good agent is rewarded based on the distance to the landmark. The adversary is rewarded if it is close to the landmark, and if the good agent is far from the landmark (the difference of the distances). Thus the adversary has to figure out how to dissuade the good agent from approaching the landmark.

The good agent observation space is comprised of the agent's own physical velocity, the relative position of the goal landmark, the color of the goal landmark, the relative positions of all the landmarks, the colors of all the landmarks, the relative positions of other agents. The adversary observation space consists of its own physical velocity, the relative positions of all landmarks and the relative position of other agents. In summary, the observation of the good agent is a $1\times 19$ vector and the observation of the adversary is a $1\times 8$ dimensional vector.

{\color{black}
In this task, our opponent model is good agent and the primary agent is the adversary. We shall investigate the learning process of our proposed DDQN-MOE for the primary agent to compete with well-trained good agent, which is obtained by ALI. It needs to note that the model of the well-trained good agent just is DDQN without MOE since we do not consider opponent modeling in ALI, but take opponent information as part of the environment.
}

\begin{figure}[H]
\label{fig:simple-push}
\centering

\subfigure[Training result for DDQN]{
\begin{minipage}[t]{0.5\linewidth}
\centering
\includegraphics[width=3in]{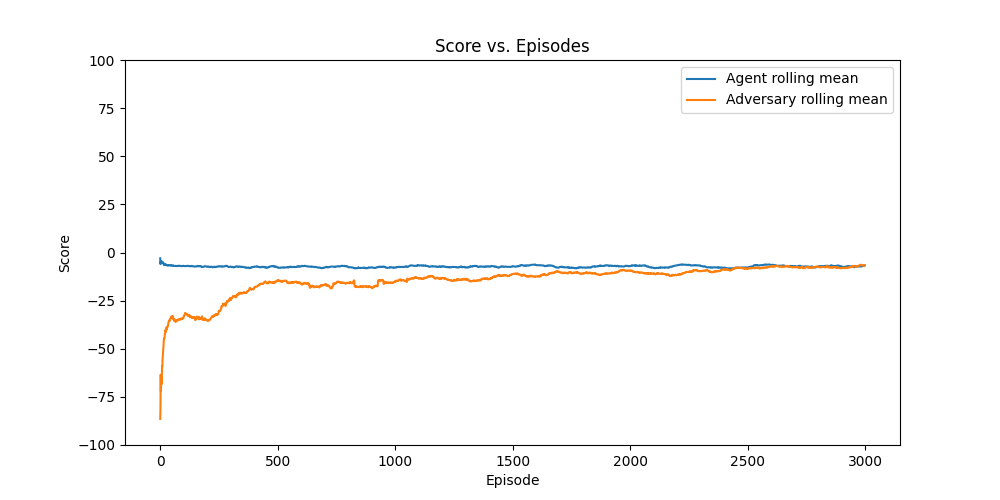}
\end{minipage}%
}%
\subfigure[Testing result for DDQN]{
\begin{minipage}[t]{0.5\linewidth}
\centering
\includegraphics[width=3in]{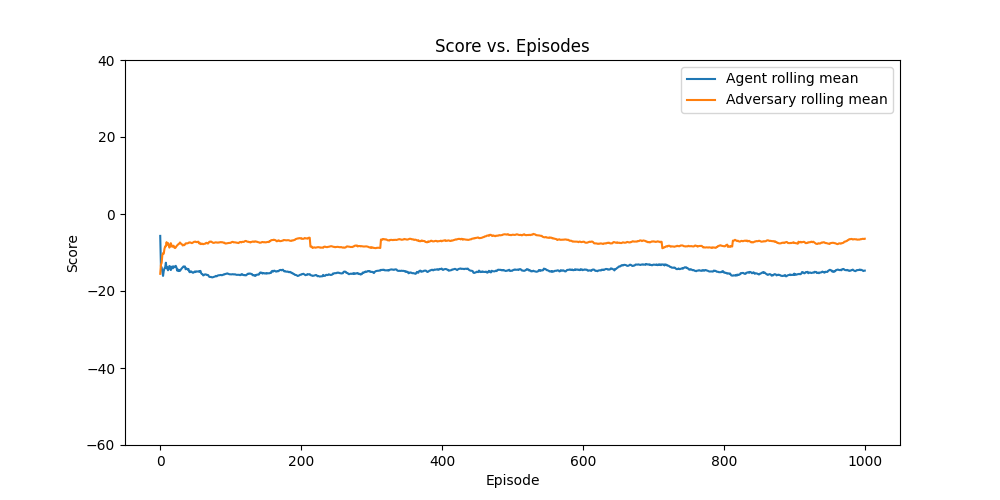}
\end{minipage}%
}%

\subfigure[Training result for DDQN-MOE]{
\begin{minipage}[t]{0.5\linewidth}
\centering
\includegraphics[width=3in]{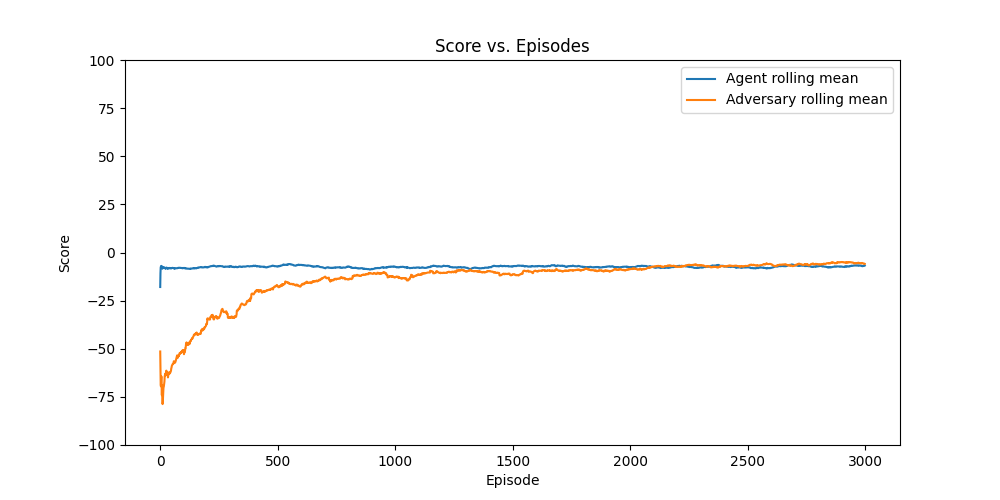}
\end{minipage}
}%
\subfigure[Testing result for DDQN-MOE]{
\begin{minipage}[t]{0.5\linewidth}
\centering
\includegraphics[width=3in]{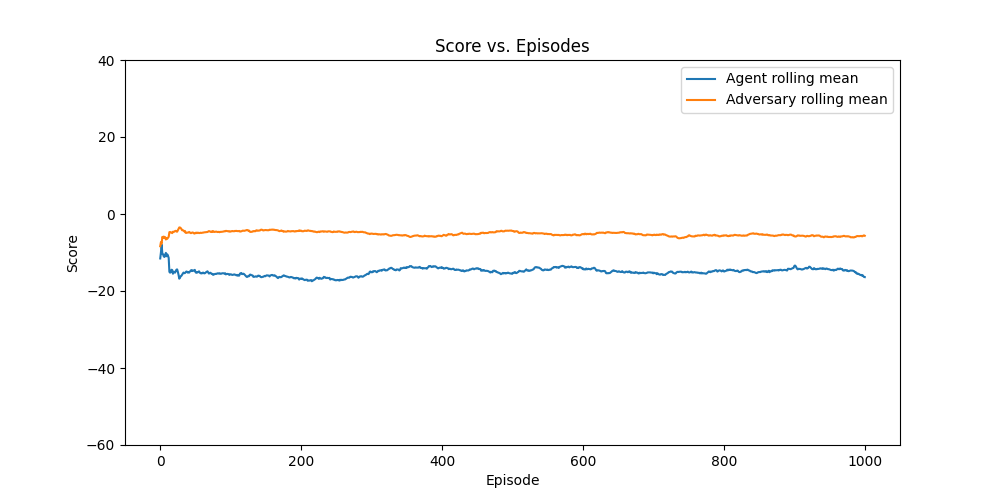}
\end{minipage}
}%
\end{figure}

{\color{black}
In Fig \ref{fig:simple-push}, the score is defined the rolling mean of the rewards within $100$ episodes. In (a) of Fig \ref{fig:simple-push}, the training process of ALI is shown, in which the models used by both agents are DDQN. And we can see that the result is an equilibrium. The test process of ALI is shown in (b), which again verifies the results after training. The random initialized adversary with DDQN-MOE model starts relearning to fight well-trained good agent with DDQN model. And the training result and testing result are presented in (c) and (b) of Fig \ref{fig:simple-push}. It can be seen that because of the introduction of MOE structure and opponent features, it reaches the equilibrium state faster.

In the testing process of (d) in Fig \ref{fig:simple-push}, the mean value and maximum value of the observed rewards for the well-trained DDQN-MOE and well-trained DDQN are presented in the Table \ref{table:simple-push}. Here Mean (agent) and Max (agent) denote the mean value and maximum value of the observed rewards for good agent. Mean (adversary) and Max (adversary) denote the mean value and maximum value of the observed rewards for good agent.

\begin{table}[H]   
\begin{center}   
\caption{Testing result for Simple Push task}  
\label{table:simple-push} 
\begin{tabular}{|c|c|c|c|c|}   
\hline   \textbf{Algorithms} & \textbf{Mean (agent)} & \textbf{Max (agent)} & \textbf{Mean (adversary)} & \textbf{Max (adversary)}\\    
\hline   DDQN-MOE & -15.1295 & -2.0557 & -5.1768 & 15.0165  \\  
\hline   DDQN  & -14.8534 & -2.4320 & -7.1575 & 13.5691 \\      
\hline   
\end{tabular}   
\end{center}   
\end{table}

In Table \ref{table:simple-push}, from the view of the primary agent adversary, we can see that the maximum and mean scores of the DDQN-MOE model based on opponent modeling are higher than those of the DDQN model without opponent modeling during the testing process, i.e., $15.0165>13.5691, -5.1768>-7.1575$. It shows that directly modeling the opponent's strategy may boost the agent's ability for learning, and it appears that this may be the case.

On the other hand, since we know that under the framework of opponent modeling, a good agent should not only obtain higher rewards, but also better limit the opponent's performance, that is, let the opponent obtain lower rewards at the same time. This is also verified in Table \ref{table:simple-push}, i.e., $-14.8534<-15.1295, -2.4320<-2.0557$.
}
\subsection{Simple Adversary}

In the Simple Adversary environment, there is 1 adversary (red), 1 good agent (green), 1 landmark.  An example of a random policy is illustrated as Fig \ref{fig:random-simple-push}.

\begin{figure}[H]
    \centering
    \includegraphics[scale=0.5]{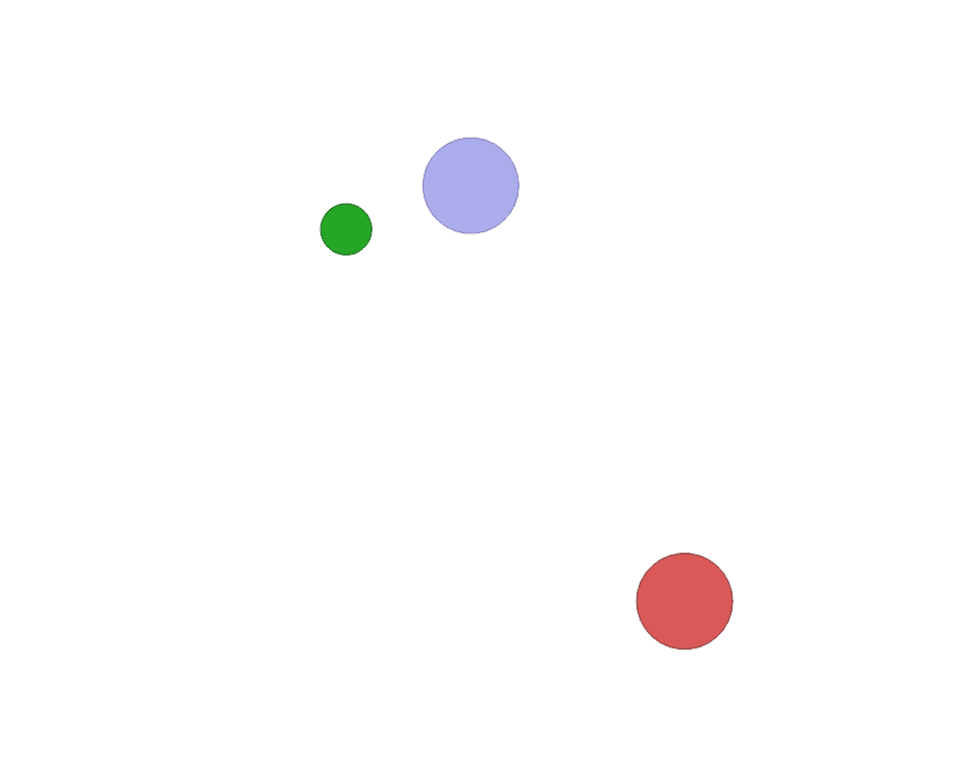}
    \caption{Random Policy on Simple Adversary}
    \label{fig:random-simple-adv}
\end{figure}

All agents observe the position of landmarks and other agents. Good agents are rewarded based on how close the closest one of them is to the target landmark, but negatively rewarded based on how close the adversary is to the target landmark. The adversary is rewarded based on distance to the target. All rewards are unscaled Euclidean distance. {\color{black} Thus the adversary must learn to push the good agent away from the landmark.
}

The agent observation space consists of its own physical position, the relative position of the goal and the relative position of other agents. The adversary observation space consists of the relative position of the landmark and the relative position of other agents. Therefore, the observation of one agent is a $1\times 6$ vector and the observation of the adversary is a $1\times 4$ dimensional vector. The action of the agent and the adversary are all $1\times 5$ vector. {\color{black} The difference between this task and the previous one is that although the purpose of this task is similar, the amount of information is smaller, so the modeling of opponent information may be more important.
}

{\color{black}
In the last task, we consider the good agent as our opponent model. As a comparison, on this task we will use the adversary as our opponent and the good agent as the primary agent. The training method and evaluation method are similar to the previous one, so we will not go into details here.
The training process and testing process for DDQN and DDQN-MOE are presented in Figure \ref{fig:simple-adv}:

\begin{figure}[H]
\label{fig:simple-adv}
\centering

\subfigure[Training result for DDQN]{
\begin{minipage}[t]{0.5\linewidth}
\centering
\includegraphics[width=3in]{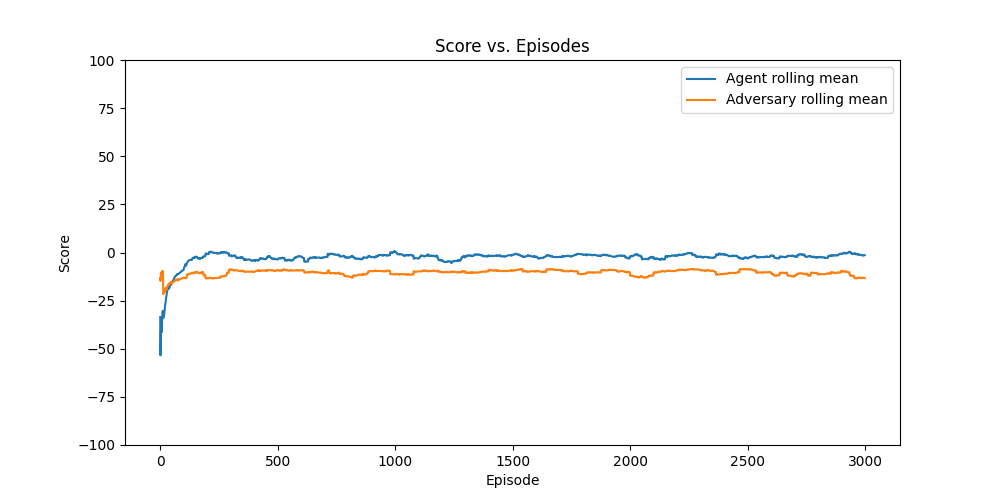}
\end{minipage}%
}%
\subfigure[Testing result for DDQN]{
\begin{minipage}[t]{0.5\linewidth}
\centering
\includegraphics[width=3in]{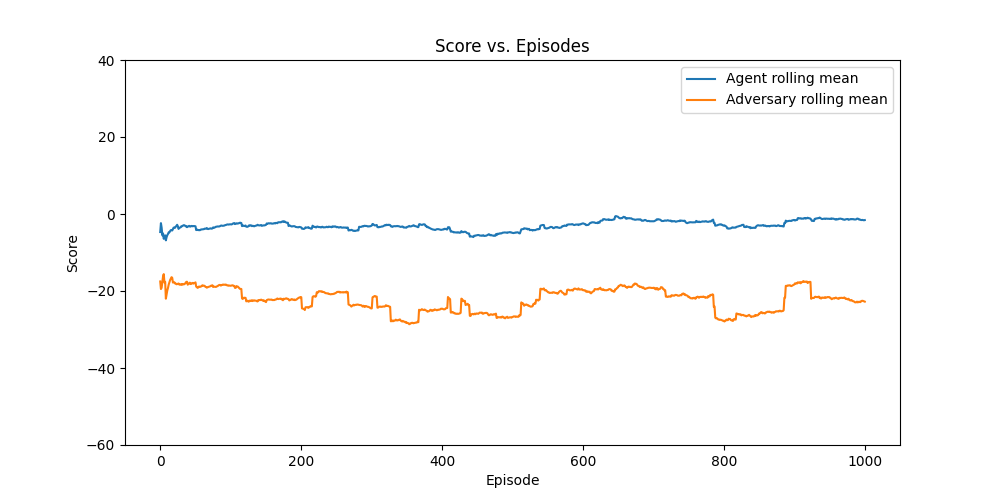}
\end{minipage}%
}%

\subfigure[Training result for DDQN-MOE]{
\begin{minipage}[t]{0.5\linewidth}
\centering
\includegraphics[width=3in]{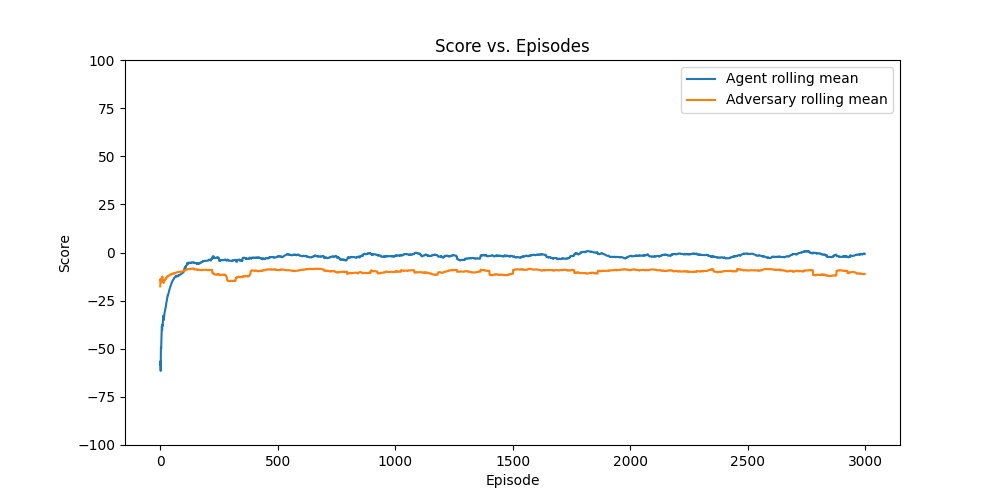}
\end{minipage}
}%
\subfigure[Testing result for DDQN-MOE]{
\begin{minipage}[t]{0.5\linewidth}
\centering
\includegraphics[width=3in]{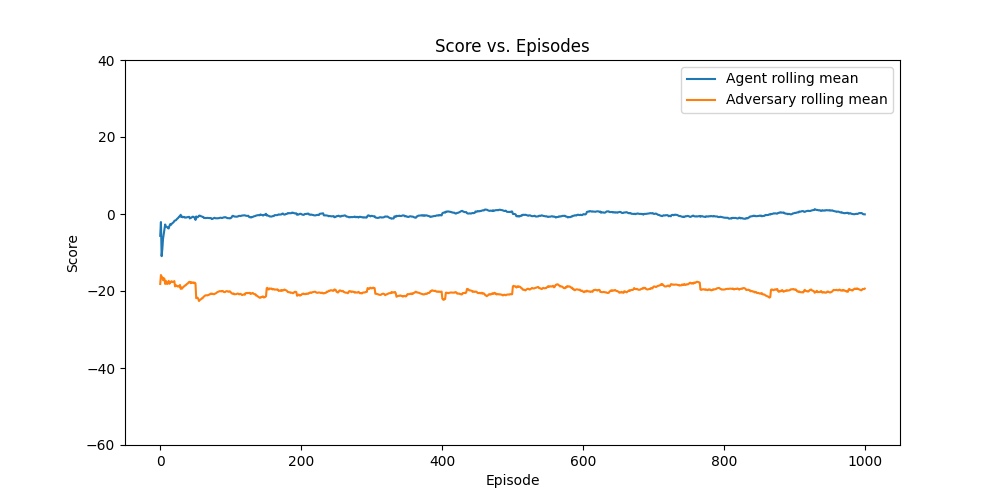}
\end{minipage}
}%
\end{figure}

From the Fig \ref{fig:simple-adv}, we see that although in the training process of (a), the equilibrium has been reached, the performance of adversary is not stable in (b). However, in testing process of DDQN-MOE, the performance of adversary is stable. This shows that our proposed DDQN-MOE is more robust compared to DDQN.

In the testing process of (d) in Fig \ref{fig:simple-adv}, the mean value and maximum value of the observed rewards for the well-trained DDQN-MOE and well-trained DDQN are presented in the Table \ref{table:simple-adv}.
\begin{table}[H]   
\begin{center}   
\caption{Testing result for Simple Adversary task}  
\label{table:simple-adv} 
\begin{tabular}{|c|c|c|c|c|}   
\hline   \textbf{Algorithms} & \textbf{Mean (agent)} & \textbf{Max (agent)} & \textbf{Mean (adversary)} & \textbf{Max (adversary)}\\    
\hline   DDQN-MOE & -0.2182 & 57.2469 & -19.9193 & -4.7093  \\  
\hline   DDQN  & -2.8690 & 16.2732 & -22.3838 & -2.2044 \\  
\hline   
\end{tabular}   
\end{center}   
\end{table}

From the Table \ref{table:simple-adv}, the conclusions similar to the last task can be drawn.
From the view of the primary agent, good agent, we can see that the maximum and mean scores of the DDQN-MOE model based on opponent modeling are significantly higher than those of the DDQN model without opponent modeling during the testing process, i.e., $-0.2182>>-2.8690, 57.2469>>16.2732$. This proves that opponent modeling is a very effective strategy for this task, and our proposed DDQN-MOE model captures and models the adversary's strategy well. 
}

\subsection{Discussions}
{\color{black}
Here we shall discuss the above experimental results. Our findings are summarized as follows:
\begin{itemize}
    \item From the above experimental results, it can be found that on these two tasks, the maximum and average scores of the DDQN-MOE model based on opponent modeling are higher than those of the DDQN model without opponent modeling during the testing process.
    
    This shows that from an algorithmic point of view, our proposed DDQN-MOE based on opponent modeling is better than the original DDQN without considering opponent modeling. Furthermore, The opponent features \eqref{eqn:opponent-features} we construct are also effective for modeling opponent strategies.
    \item Our proposed DDQN-MOE not only achieves higher scores, but also lowers the opponent's score, demonstrating that our model delivers a superior learning technique.  And DDQN-MOE learnt more discriminative opponent representation helps model the relation between opponent behavior and Q-values.
    \item We discover that the good agent and the adversary engage in competition throughout training without MOE and eventually reach an state of equilibrium. The inclusion of opponent modeling and the MOE model disrupts this equilibrium and improves the performance of the agent with opponent features. This is actually a fairly natural conclusion, because it breaks the imbalance of information, and we point it out here just to show that opponent modeling is just a way to break this equilibrium.
\end{itemize}

}

\section{Conclusion and Summary}
In this study, we propose the DDQN-MOE model, which incorporates Mixture-of-Experts architecture into an opponent modeling framework that makes use of Double Deep Q Network and a prioritized experience replay mechanism. When we evaluated our model in two different multi-agent situations and compared its performance to that of DDQN, we found that our model performed much better than DDQN's. In the future, research may concentrate on developing more accurate models of the strategies and traits of potential adversaries.

\begin{appendices}\label{appendix-params}
  \section*{The Parameters in Algorithm Implementation}
  The deep Q-networks used in this paper are all feed-forward neural network with two hidden layers using Rectified Linear Units (ReLU) \cite{Agarap2018DeepLU} activation function. The units of the first hidden layer is $64$ and the units of the second hidden layer is $128$. The Q-networks for the good agent and adversary have the same hidden units but with different input units. The detailed hyperparameters about the training process for these two tasks are shown in the following Tables.
\begin{table}[H]
\label{table:simple-push-param}
\caption{DDQN Hyperparameter Setup for Simple Push}
\begin{center}
\begin{tabular}{cc}
\toprule
Hyperparameter & Value  \\
\midrule
minibatch size & 64 \\
replay memory size & 1,000,000 \\
target network update frequency & 5 \\
discount factor & 0.999 \\
learning rate & 0.0001 \\
initial $\epsilon$ & 1.0 \\
final $\epsilon$ & 0.1 \\
replay start size &  50000 \\
total training episodes & 3000 \\
total testing episodes & 1000 \\
loss function  & MSE \\
optimizer & Adam \\
$\alpha$ & 0.6 \\
$\beta$ & 0.4 \\
$\epsilon$ (PER) & 0.00001 \\
annealing steps & 1000000 \\
\bottomrule
\end{tabular}
\end{center}
\end{table}

\begin{table}[H]
\label{table:simple-adversary-param}
\caption{DDQN Hyperparameter Setup for Simple Adversary}
\begin{center}
\begin{tabular}{cc}
\toprule
Hyperparameter & Value  \\
\midrule
minibatch size & 64 \\
replay memory size & 100000 \\
target network update frequency & 5 \\
discount factor & 0.999 \\
learning rate & 0.001 \\
initial $\epsilon$ & 1.0 \\
final $\epsilon$ & 0.1 \\
replay start size &  50000 \\
total training episodes & 3000 \\
total testing episodes & 1000 \\
loss function  & MSE \\
optimizer & Adam \\
$\alpha$ & 0.6 \\
$\beta$ & 0.4 \\
$\epsilon$ (PER) & 0.00001 \\
annealing steps & 1000000 \\
\bottomrule
\end{tabular}
\end{center}
\end{table}
\end{appendices}

\bibliographystyle{apacite}
\bibliography{sample}

\end{document}